# Dynamic Sliding Mode Control based on Fractional Calculus Subject to Uncertain Delay Based Chaotic Pneumatic Robot

Sara Gholipour P.[1,2,3,4], Heydar Toosian Sh.[4], Mobin Alizadeh[1,2,3], Sara Minagar[1,2,3] and Seyed Javad Kazemitabar[3,*]

[1]Farabina Noshirvani's smart Co.
[2]Robotic Research Lab., Babol Noshirvani Unviersity of Technology
[3]Faculty of Electrical and Computer Engineering, Babol Noshirvani Unviersity of Technology, Babol, Iran
[4]Faculty of Electrical and Robotic Engineering, Shahrood University of Technology, Shahrood, Iran

A B S T R A C T

This paper considers the chattering problem of sliding mode control in presence of delay in robot manipulator causing chaos in such electromechanical systems. Fractional calculus was used in order to produce a novel sliding mode to eliminate chatter. To realize the control of a class of chaotic systems in master-slave configuration, a novel fractional dynamic sliding mode control scheme is presented and examined on the delay based chaotic robot. Also, the stability of the closed-loop system is guaranteed by Lyapunov stability theory. Moreover, delayed robot motions are sorted out for qualitative and quantitative study. Finally, numerical simulations illustrate feasibility of the proposed control method.

## 1. Introduction

Due to inherent insensitivity characteristic of Sliding Mode Control (SMC) [1, 2] an undesirable oscillation known as chattering occurs that affects the dynamical system. In order to attenuate the chattering some techniques such as boundary layer, fuzzy, observer, relay control gain and etc. have been proposed [3-6]. However, these methods do not confirm sliding mode condition exactly. Dynamic SMC (DSMC) [7-12], a generalized controller of SMC, is a powerful chattering reduction technique and is based on injecting dynamics to the major system. Fractional DSMC (FDSMC) which is established on fractional derivatives, is propounded for a class of chaotic systems in a master-slave configuration and particularly presented on a robot manipulator when delay exists as an inseparable content of actual systems leading to chaotic motions. Fractional derivatives [13-17] were well examined in many controllers either in theoretical or industrial applications [18-23]. They are also being considered in system modeling [24, 25]. Chaos [26-31] appearance in 2-link robot manipulators with their important role in many different situations, can be achieved by time delays, external inputs, system parameters and controllers [32-35]. To realize task and joint space of chaotic dead time 2-link robot motions [32] and qualitative and quantitative characteristics such as phase attractor, time series, maximum Lyapunov exponent, bifurcation diagram with respect to dead time and Poincaré map are investigated. Finally, the proposed fractional controller eventuates chattering elimination and tracks the same periodic 2-link manipulator in a finite time even in presence of high uncertainty. This rest of this paper is organized as follows: next section prepare some preliminaries of fractional calculus. In Section 3 our proposed fractional dynamic sliding mode controller is presented. The dynamics of the two-link robot and their control method is studies in Section 4. Moreover, numerical simulation results are presented in Section 5 to show the effectiveness of the proposed method. Finally, in Section 6 the paper is concluded.

---
[*] *Corresponding author.*
 *E-mail addresses:* j.kazemitabar@nit.ac.ir

## 2. Preliminaries

We devote this section to basic definition of fractional calculus and its properties (Table 1). The Riemann–Liouville (RL) and the Caputo are the most popular fractional derivatives definitions.

*RL definition:*

$$_aD_t^q f(t) = \frac{1}{\Gamma(n-q)} \frac{d^n}{dt^n} \int_a^t \frac{f(\tau)}{(t-\tau)^{q-n+1}} d\tau, \tag{1}$$

*Caputo definition:*

$$_aD_t^q f(t) = \frac{1}{\Gamma(n-q)} \int_a^t \frac{f^{(n)}(\tau)}{(t-\tau)^{q-n+1}} d\tau, \tag{2}$$

Where $_aD_t^q$ denotes the $q^{th}$ order of differintegral operator, $a$ and $t$ are the limits of operation, $\Gamma(x)$ is the well-known Gamma function and $n$ is the first integer not less than $q$, i.e. $n-1 < q < n$.

**Table 1**
Some useful properties of fractional definitions

| | Caputo Definition | Riemann-Liouville Definition |
|---|---|---|
| 1 | $_a^C D_t^\lambda \frac{d^m}{dt^m} f(t) = {_a^C D_t^{m+\lambda}} f(t) = \frac{d^m}{dt^m} {_a^C D_t^\lambda} f(t)$ <br> only if $f^{(k)}(0) = 0$, $k = n, n+1, ..., m$ <br> ($m = 0, 1, 2, ...; n-1 < \lambda < n$) | $\frac{d^m}{dt^m} {_a^{RL} D_t^\lambda} f(t) = {_a^{RL} D_t^{m+\lambda}} f(t) = {_a^{RL} D_t^\lambda} \frac{d^m}{dt^m} f(t)$ <br> only if $f^{(k)}(0) = 0$, $k = 0, 1, 2, ..., m$ <br> ($m = 0, 1, 2, ...; n-1 < \lambda < n$) |
| 2 | $_a^C D_t^\alpha {_a^C D_t^\beta} f(t) = {_a^C D_t^{\alpha+\beta}} f(t) = {_a^C D_t^\beta} {_a^C D_t^\alpha} f(t), \alpha \neq \beta$ <br> if $0 < \alpha, \beta < 1, \alpha + \beta \in (0,1]$ | $_a^{RL} D_t^\alpha {_a^{RL} D_t^\beta} f(t) = {_a^{RL} D_t^{\alpha+\beta}} f(t) = {_a^{RL} D_t^\beta} {_a^{RL} D_t^\alpha} f(t), \alpha \neq \beta$ <br> if $0 < \alpha, \beta < 1, \alpha + \beta \in (0,1]$ |
| 3 | $_0^C D_t^\alpha A = 0$, $A = cte$ | $_0^{RL} D_t^\alpha A = \frac{t^{-\alpha}}{\Gamma(1-\alpha)} A$, $A = cte$ |

## 3. Fractional Dynamic Sliding Mode Control

### 3.1 System description

Consider a MIMO nonlinear differential system described by:

$$\begin{aligned} \dot{x} &= f(t,x,u) + \Delta f(t,x,u) + d_0(t) \\ z &= h(t,x,u) \end{aligned} \tag{3}$$

Where $x \in \mathbb{R}^n$ is the state vector, $u \in \mathbb{R}^m$ is a control input vector, $y \in \mathbb{R}^p$ is the output vector and $f(t,x,u), g(t,x,u)$ are smooth nonlinear functions.

We define

$$d(t,x,u) = \Delta f(t,x,u) + d_0(t) \tag{4}$$

as a continuous-differential and $d(t,x,u), \dot{d}(t,x,u)$ are limited functions.

### 3.2 Design of Fractional Dynamic Sliding Mode Control (FDSMC)

Consider the Eq. (3) as a following master system:

$$x^{(n)} = f(x,t) \tag{5}$$

And the following slave system with input controls:

$$y^{(n)} = g(y,t) + b(y,t)u \tag{6}$$

Where

$$\begin{aligned} &0 < b_{min} \leq b \leq b_{max}, \quad 0 < a_{min} \leq \dot{b} \leq a_{max}, \\ &\hat{b} = (b_{min} b_{max})^{1/2}, \quad \beta^{-1} \leq \hat{b} b^{-1} \leq \beta, \quad \beta = (b_{max}/b_{min})^{1/2} \\ &|\Delta f| < f_1, \ |\Delta \dot{f}| < f_2, \ |\Delta g| < g_1, \ |\Delta \dot{g}| < g_2. \end{aligned} \tag{7}$$

and the tracking error is:
$$e = y - x \tag{8}$$

**First Step** (Existence Problem) designing a switching function which provides the desirable system performance:

$$\begin{aligned}S &= s_1(t,\dot{e},D^\lambda \dot{e},...,D^{(n+\lambda-1)}\dot{e},u) \\ &= k_0\dot{e} + k_1 D^\lambda \dot{e} + k_2 D^{\lambda+1}\dot{e} + ... + k_n D^{n+\lambda-1}\dot{e}\end{aligned} \tag{9}$$

where $k_i > 0$, $i = 1, 2, ..., n$

**Second Step** (Reachability Problem) consists of obtaining a control law that the first order derivative of the control input is derived from the system states and also control input, so the dynamics of the system are added. Taking the time derivative of Eq. (9), we obtain:

$$\begin{aligned}\dot{S} &= s_2(t,\ddot{e},\frac{d}{dt}D^\lambda \dot{e},...,D^{n+\lambda-1}\dot{e},\dot{u}) \\ &= k_0\ddot{e} + k_1 \frac{d}{dt}D^\lambda \dot{e} + k_2 \frac{d}{dt}D^{\lambda+1}\dot{e} + ... + k_n \frac{d}{dt}D^{n+\lambda-1}\dot{e}\end{aligned} \tag{10}$$

**Lemma:** The motion of the sliding mode is asymptotically stable, if the following condition is held $\dot{S}S < 0$.

**Proof:** Consider the following Lyapunov candidate function:
$$V = \frac{1}{2}S^2 \tag{11}$$

by the following condition:
$$\dot{S} = -K_s \operatorname{sgn}(S) \tag{12}$$

where **sgn**(S) is signum function:
$$\operatorname{sgn}(S) = \begin{cases} +1, & \text{if } S > 0, \\ 0, & \text{if } S = 0, \\ -1, & \text{if } S < 0. \end{cases} \tag{13}$$
(14)

The time derivative of Eq.(11) is: $\dot{V} = \dot{S}S = -K_s S.\operatorname{sgn}(S)$ since $S.\operatorname{sgn}(S) > 0$ and $K_s > 0$ we have $\dot{V} = \dot{S}S < 0$, hence $\dot{V}$ is negative definite, thus $S(t)$ is toward the switching surface and sliding mode is asymptotically stable. □

The controller is then solved from $\dot{S} = -K_s.\operatorname{sgn}(S)$ First order derivative of the control law is as follows:

$$\begin{aligned}\dot{u} &= h(t,\dot{e},...,D^\alpha e,u) \\ &= -\hat{b}^{-1}(y,t)\{\frac{k_0}{k_n}D^{1-\lambda}\dot{e} + \frac{k_1}{k_n}\ddot{e} + ... \\ &+ \dot{\hat{b}}\hat{b}^{-1}(\dot{e} - \hat{g}(y,t) + \hat{f}) + \dot{\hat{g}}(y,t) - \dot{\hat{f}}(x,t) + K_s D^{-\lambda}\operatorname{sgn}(S)\}\end{aligned} \tag{14}$$

If the following control gain is selected by this sliding surface condition $\dot{S}.\operatorname{sgn}(S) \leq -\eta$ where $\eta$ is a positive real constant, the stability of the system is satisfied:

$$\begin{aligned}K_s &\geq \beta(k_2^{-1}\eta + (1 - \beta^{-1})U \\ &+ (\dot{b}\hat{b}^{-1} - k_2\dot{\hat{b}}\hat{b}^{-1})e^{(n)} + k_2(g_1 + g_2 + f_1 + f_2))\end{aligned} \tag{15}$$

Suppose that:
$$\begin{aligned}\dot{\hat{u}} &= -\frac{k_0}{k_n}D^{1-\lambda}\dot{e} - \frac{k_1}{k_n}\ddot{e} + ... - \dot{\hat{b}}\hat{b}^{-1}e^{(n)}, \quad |\dot{\hat{u}}| < U \\ \Delta f &= (\hat{b} - b)\hat{b}^{-1}\hat{f}, \quad \Delta \dot{f} = b\hat{b}^{-1}\dot{\hat{f}} - \dot{f}, \\ \Delta g &= (-\dot{b}\hat{b}^{-1} + b\hat{b}^{-1}\dot{\hat{b}}\hat{b}^{-1})\hat{g}, \quad \Delta \dot{g} = \dot{g} - b\hat{b}^{-1}\dot{\hat{g}}\end{aligned} \tag{16}$$

Finally, control law is obtained as integral of Eq. (14).

## 4. Application to control

*4.1 Information about two-link pneumatic robot with a dead time*

Pneumatic robots serve a wide variety of industries, but long air tube which connects actuator and transducers of pneumatic robots causes a dead time in control section that even includes chaotic motion. Fig. 1. shows a) a two-link pneumatic robot and b) its block diagram with PD control, where $\theta_d(t) = \frac{\pi}{4}\sin(0.5\pi t)$ is the desired value and the PD controller is: $K_p + K_d s$ ($K_p = K_d = 4$). The dynamics of the system is presented as [33]:

$$M(\theta(t))\ddot{\theta}(t) + H(\theta(t),\dot{\theta}(t)) + D\dot{\theta}(t) = \tau(t - L) \tag{17}$$

where $\theta(t)$ is the joint angle, $M(\theta(t))$ is the inertia matrix, $H(\theta(t),\dot{\theta}(t))$ is the centripetal and Coriolis torque, $D$ is the viscous friction and $\tau$ is input torque with a dead time $L$.

Here are each element expressions:

$$\theta(t) = [\theta_1(t), \theta_2(t)]^T \tag{18}$$

$$M(\theta(t)) = \begin{bmatrix} J_1 + J_2 + 2\beta\cos\theta_2(t) & J_2 + \beta\cos\theta_2(t) \\ J_2 + \beta\cos\theta_2(t) & J_2 \end{bmatrix} \tag{19}$$

$$H(\theta(t),\dot{\theta}(t)) = \begin{bmatrix} -\beta(2\dot{\theta}_1(t)\dot{\theta}_2(t) + \dot{\theta}_2(t)^2)\sin\theta_2(t) \\ \beta\dot{\theta}_1(t)^2\sin\theta_2(t) \end{bmatrix} \tag{20}$$

$$D = \begin{bmatrix} D_1 & 0 \\ 0 & D_2 \end{bmatrix}, \quad \tau(t-L) = \begin{bmatrix} \tau_1(t-L) \\ \tau_2(t-L) \end{bmatrix} \tag{21}$$

$$J_1 = I_1 + (m_1 + 4m_2)l_1^2, \quad J_2 = I_2 + m_2 l_2^2,$$
$$\beta = 2m_2 l_1 l_2, \quad I_1 = \frac{1}{3}m_1 l_1^2, \quad I_2 = \frac{1}{3}m_2 l_2^2. \tag{22}$$
$$D_1 = D_2 = 0.5 Nms, \, l_1 = l_2 = 0.25m, \, m_1 = m_2 = 1.0kg$$

where $2l_i$, $m_i$ are length and mass of link $i$($i$=1,2), respectively. We have investigated some characteristics of the plant such as the end effector motion in workspace, time series, bifurcation diagram with respect to delay parameter and reconstitute attractor, the attractor and Poincaré map that are shown in Figs. 2-4. Except bifurcation diagram, time delay is $L$=0.015 in all of above figures and the maximum lyapunov exponent is equal to *0.041*.

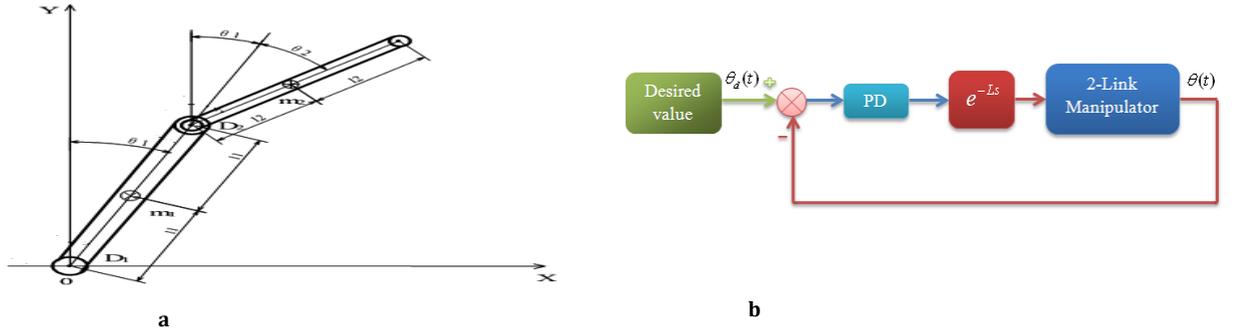

**Fig. 1.** a) A two-link pneumatic robot and b) its block diagram with PD control

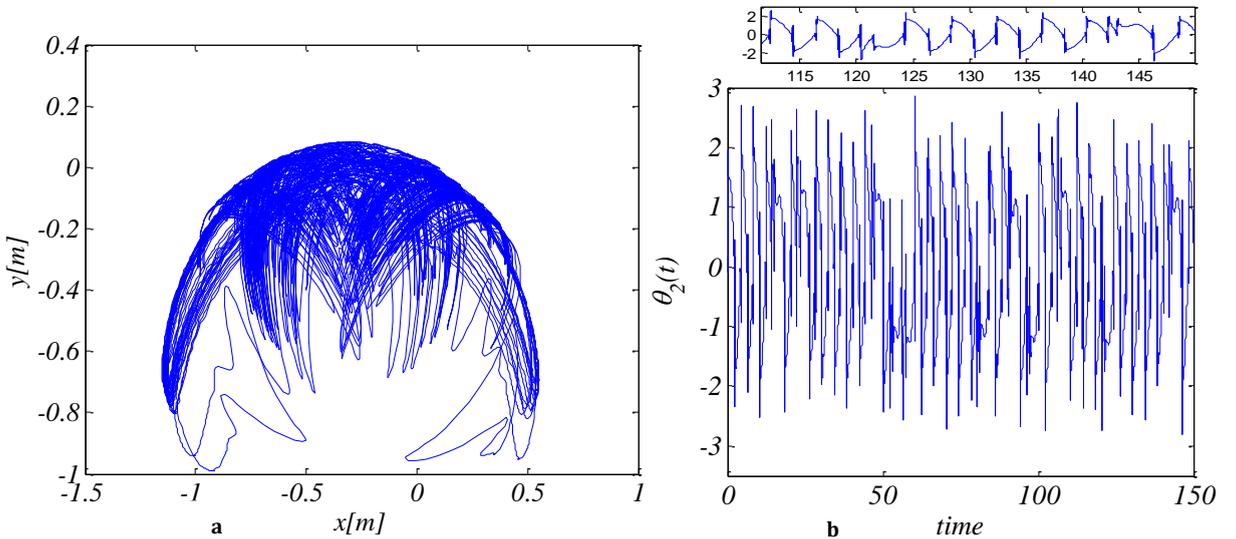

**Fig. 2.** a) The end-effector motion in workspace and b) Time series of link2 in joint-space; chaotic pneumatic robot with L=0.015[sec]

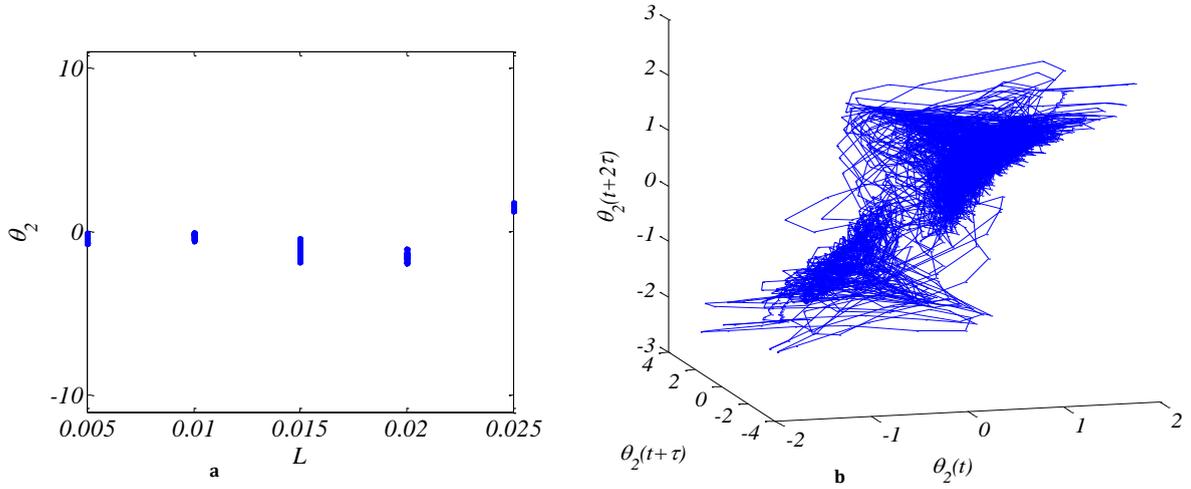

**Fig. 3.** a) Bifurcation diagram w.r.t $L$ and b) Reconstitute attractor of link2 with $\tau = 5$ [sec] of chaotic pneumatic robot ($L=0.015[sec]$)

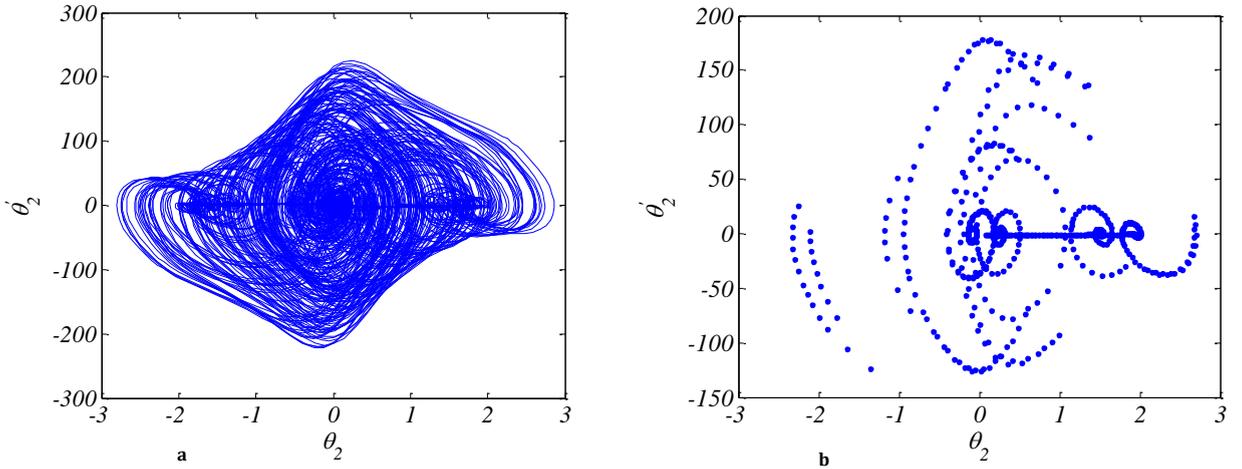

**Fig. 4.** a) The attractor in joint-space and b) Poincaré map on plane $\theta_1 = 0.5\,[rad]$; chaotic pneumatic robot with $L=0.015[sec]$

*4.2 Control of plant with uncertainty*

We consider control between two pneumatic robots by different time delays demonstrated by considering the system defined in Eq. (17) with *L=0.005[sec]* as the master system, and the slave system with *L=0.015[sec]*.

**Theorem 2:** If the control scheme satisfies:
$$\dot{T}_s = M_s \{ \frac{K_p}{K_f} D^{1-\lambda}\dot{e} + \frac{K_d}{K_f}\ddot{e} + \theta_m^{(3)} \\ - \dot{M}_s^{-1}(T_s - h_s) + M_s^{-1}\dot{h}_s \} + k_s . D^{-\lambda} sign(S) \tag{23}$$

The error will be asymptotically stable, where *m* denotes master and *s* denotes slave and
$$T(t) = \tau(t - L).$$

*Proof:* if we define the following surface:
$$\dot{S} = K_p \ddot{e} + K_d \frac{d}{dt} D^\lambda \dot{e} + K_f \frac{d}{dt} D^{\lambda+1}\dot{e} \tag{24}$$

Using the method in section 3 the dynamic control scheme results as (24).□

In next section results with and without maximum uncertainty of 60 percent in slave system are exhibited.

## 5. Simulation results

Simulation results are shown in Figs. 5-9. Using MATLAB software and the following parameter values:
$$K_s = 1, K_p = 2, K_d = 10, K_f = 0.1 \text{ and } \lambda = 0.7 \tag{25}$$

To evaluate the robustness of the control scheme, a system uncertainty term is inserted by maximum uncertainty of 60 percent in slave system:

$l_1 = l_2 = 0.15m, m_2 = 0.4kg, m_1 = 0.7kg$ (26)

Figs. 5-7 and Figs. 8-9 show results of control of two pneumatic robots with and without uncertainty, respectively (The end-effector motion in workspace, tracking error of link(2), sliding surfaces $S$ and control inputs $u$, the attractors and reconstitute attractor) via fractional dynamic sliding mode control (activated in t=0.1sec). Numerical simulations show torus behavior in a finite time and chattering free by root-mean-square error of link (2) with and without uncertainty are 7.089e-5, 1.675e-2, respectively and maximum Lyapunov exponent is equal to *0.0032.*

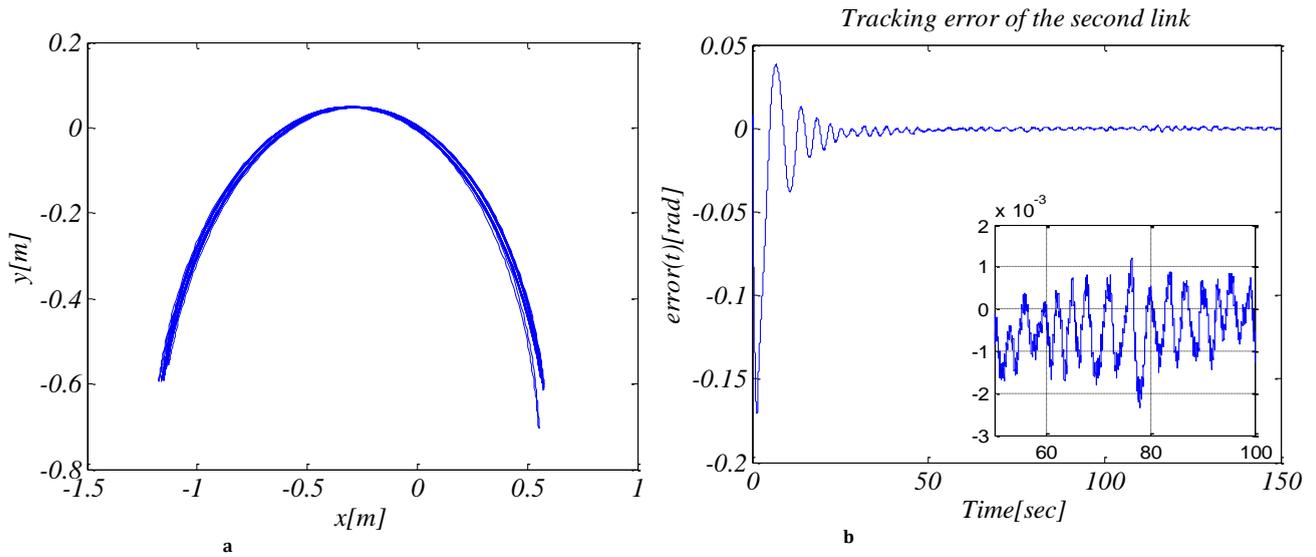

**Fig. 5.** Result of control between two pneumatic robot via fractional dynamic sliding mode control (activated in t=0.1[sec]);
a)   The end-effector motion in workspace and b) Tracking error of link(2)

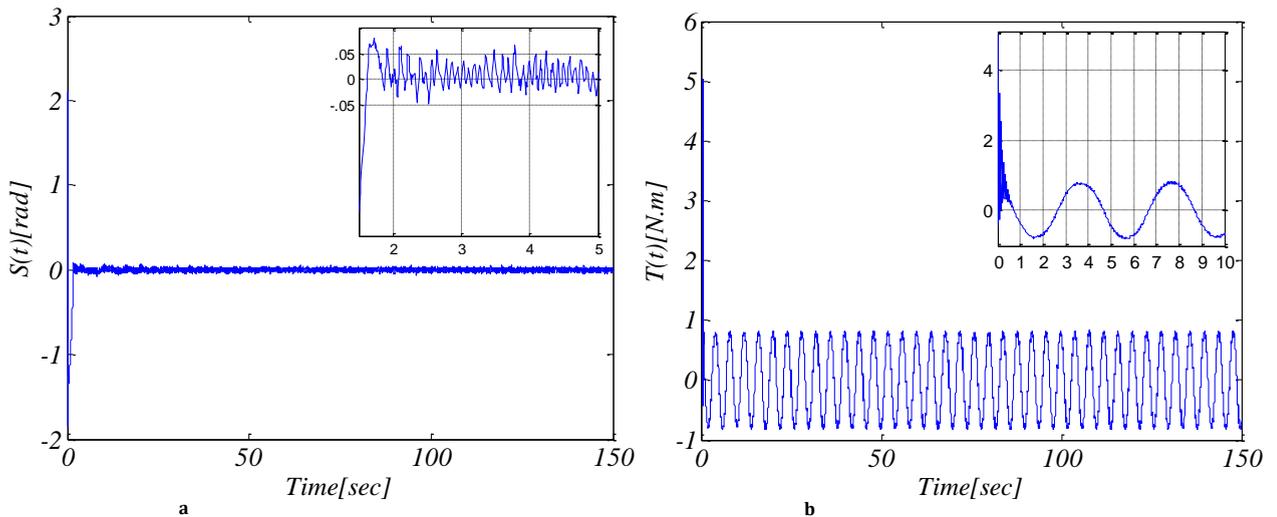

**Fig. 6.** Result of (Sliding surfaces and Control inputs) control between two pneumatic robots via fractional dynamic sliding mode control (activated in t=0.1[sec])

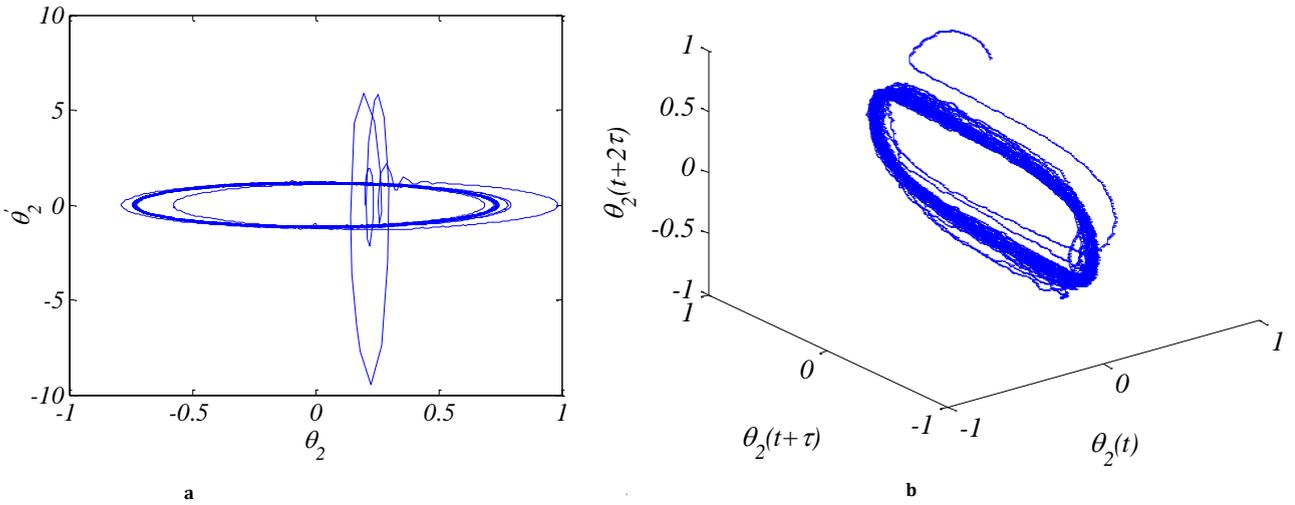

**Fig. 7.** Results of control between two pneumatic robots via fractional dynamic sliding mode control (activated in t=0.1[sec])
a) The attractor and b) Reconstitute attractor with

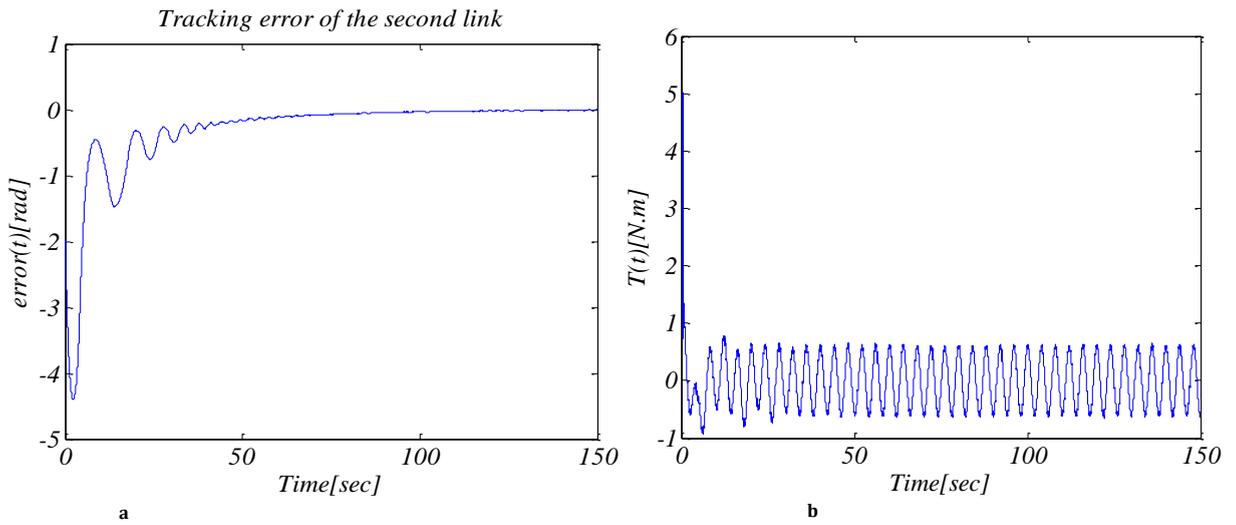

**Fig. 8.** Results of control between two pneumatic robots with uncertainty via fractional dynamic sliding mode control (activated in t=0.1[sec])
a) Tracking error of link(2) b) Control input

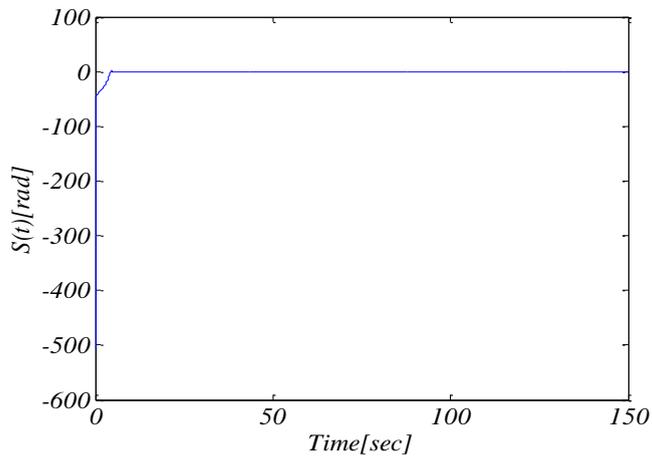

**Fig. 9.** Result of (sliding surface) control between two pneumatic robot with uncertainty via fractional dynamic sliding mode control (activated in t=0.1[sec])

# 6. Conclusion

Chattering problem and particularly a delay based chaotic pneumatic robot has been discussed. Fractional dynamic sliding mode has been proposed and examined in a chaotic robot system in presence of maximum uncertainty of 60 percent. Numerical examples indicated the viability of control scheme.